\documentclass[oneside,10pt]{swissi}

\newcommand{\papertitle}{Functional Architecture of European Electricity Trading Markets}
\newcommand{\papersubtitle}{Requirements for AI Supported Trading Systems under Regulatory Constraints}

\newcommand{\papercitationauthors}{Kurz, W. and Stricker, W.}

\newcommand{\paperauthors}{%
  Walter Kurz\textsuperscript{1}%
 Wojtek Stricker\textsuperscript{1}%
}

\newcommand{\paperaffiliations}{%
  \textsuperscript{1}Swissi Institute for AI,
  \href{mailto:kurz@swissi-ai.institute}{kurz@swissi-ai.institute};
  \href{mailto:stricker@swissi-ai.institute}{stricker@swissi-ai.institute}%
}

\newcommand{\paperdeclarationaffiliations}{%
  \begin{tabular}[t]{@{}l@{}}
    \textsuperscript{1}Swissi Institute for AI,
    \href{mailto:kurz@swissi-ai.institute}{kurz@swissi-ai.institute};
    \href{mailto:stricker@swissi-ai.institute}{stricker@swissi-ai.institute}%
  \end{tabular}%
}

\newcommand{\papercorresponding}{%
  Walter Kurz,
  \href{mailto:kurz@swissi-ai.institute}{kurz@swissi-ai.institute},
  \href{https://orcid.org/0009-0006-8045-4775}{ORCID 0009-0006-8045-4775}%
}

\newcommand{\paperfunding}{%
  This research received no external funding.%
}
\newcommand{\paperconflictsofinterest}{%
  The authors declare no conflicts of interest.%
}
\newcommand{\paperdataavailability}{%
  The data and code supporting this article are available from the corresponding author on reasonable request.%
}
\newcommand{\paperaitools}{%
  Generative AI tools were used for editorial work only, such as language editing and formatting, in accordance with international scientific standards. The authors verified the content and remain fully responsible for the article.%
}
\newcommand{\paperauthorcontributions}{%
  All authors contributed equally to this article.%
}

\title{\papertitle \\[0.35em]
\normalsize \papersubtitle}

\author{{\fontsize{8}{10}\selectfont\paperauthors}\\[0.9em]
{\scriptsize\paperaffiliations}}
\date{\vspace{1.5em}}

\begin{document}
\maketitle

\begin{abstract}
European electricity trading in the EU operates as a constrained multi-layer system in which legal design, exchange microstructure, and network physics are executed jointly across forward, day-ahead, intraday, and balancing horizons. This paper develops a functional architecture for AI-supported trading that is aligned with market-coupling mechanics, cross-zonal transfer constraints, and compliance obligations under REMIT, MiFID II, MiFIR, and EMIR. The contribution is a formal system specification composed of a decision-state vector, residual-exposure accounting, constrained optimization objective, executable-action permission gate, and fail-closed AI control logic with auditable records. The analysis maps major Nominated Electricity Market Operator (NEMO) venues and related exchange operators into an operational venue topology and identifies where cross-border coordination fails in practice: interface-level timing, permission heterogeneity, and balancing-layer coupling. The resulting framework proposes how AI can be deployed as a bounded decision component inside regulated market operation with explicit governance, rather than as an unconstrained prediction layer.
\end{abstract}


In simple terms, energy trading starts long before physical delivery: developers, utilities, and investors decide which assets to build (for example wind, solar, flexible thermal units, and batteries) by estimating future revenue across wholesale markets and system services. Once assets are online, producers and retailers hedge part of their expected volume ahead of delivery, then set the main hourly position in the day-ahead auction, where clearing reflects both bids and available cross-border transmission capacity. After day-ahead results are published, forecast errors from weather, demand, or outages create position gaps, and market participants rebalance in intraday trading where prices can move quickly as new information arrives. Arbitrage appears when the same megawatt-hour has different prices across hours or bidding zones, so traders buy in lower-priced contexts and sell in higher-priced contexts, limited by transfer capacity and timing constraints. Batteries are important in this chain because they move energy through time: charge in low or negative price periods, discharge in high-price periods, and monetize short-cycle volatility that inflexible assets cannot capture. Close to real time, TSOs activate balancing energy to keep system frequency stable, and remaining deviations are settled financially as imbalance cost. From an investor viewpoint, the full business case is an end-to-end stack in which project development quality determines optionality, trading quality determines spread capture, and governance quality determines how much gross margin survives collateral, penalties, and settlement friction.\cite{ec_electricity_market_design_2026,acer_market_coupling_development_2026,nord_pool_day_ahead_2026,nord_pool_intraday_2026,eu_reg_2017_2195_ebgl}

\section{Introduction}
European electricity trading is a constrained coordination problem in which legal rules and system physics are executed jointly. Orders clear under cross-zonal transfer limits, bidding-zone definitions, and balancing responsibility assignments, so dispatch-feasible outcomes depend on both network conditions and market design parameters.\cite{eu_reg_2019_943,acer_market_coupling_development_2026}

EU operation is distributed across linked horizons: forward capacity rights, day-ahead coupling, intraday re-dispatch, and balancing activation close to delivery. These horizons are not independent layers; positions transferred from one horizon constrain feasible actions in the next, and coupling algorithms apply common cross-zonal capacity inputs at gate closure.\cite{eu_reg_2016_1719,acer_market_coupling_development_2026}

Exchange plurality creates a second constraint class. Member States designate one or more Nominated Electricity Market Operators, and designated operators can provide services across borders under passporting conditions while market coupling operator functions are executed jointly.\cite{acer_market_coupling_development_2026} Trading systems that operate across regions must align venue-specific access rules, order handling, and fallback procedures with synchronized coupling timelines, which creates implementation friction even when legal harmonization exists.\cite{acer_final_assessment_report_2022}

The policy baseline shifted after the 2021--2022 crisis period. The reform package that entered into force on 16 July 2024 kept marginal pricing and strengthened long-term contracting and consumer-risk protection tools.\cite{ec_electricity_market_design_2026} ACER assessments report substantial welfare gains from integrated market operation and identify unresolved constraints in cross-border capacity availability and implementation quality.\cite{acer_final_assessment_report_2022}

The current AI literature base in electricity markets is still concentrated on sub-task classes. Reviews synthesize forecasting-model performance for day-ahead, intraday, and balancing price prediction, while reinforcement-learning studies optimize bid construction mainly for day-ahead auction settings.\cite{lago_marcjasz_schutter_weron_2021_epf,oconnor_bahloul_prestwich_visentin_2025_epf,di_persio_garbelli_giordano_2025_rl_bidding} These strands improve local components but do not provide an integrated architecture that jointly binds cross-zonal coupling constraints, multi-venue permissions, and compliance-by-design controls across end-to-end execution.

The missing element is an operational architecture that maps legal obligations, market-coupling mechanics, and actor permissions into implementable system requirements for AI-supported trading. This paper proposes such an architecture through five research questions: how horizon layers interact as one functional system, which actors control binding decisions, which governance constraints define executable permissions, which state variables are required for decision-ready operation, and how AI support can improve decisions while preserving auditability and compliance.

\section{Market design}
European electricity market design is a synchronized multi-horizon control system rather than a sequence of independent auctions. Legal design in Regulation (EU) 2019/943 and the network-code stack couples long-term allocation, day-ahead clearing, intraday correction, and balancing settlement through common consistency requirements on cross-zonal capacity use and balancing responsibility.\cite{eu_reg_2019_943,ec_network_codes_2026,eu_reg_2016_1719}

The forward layer allocates transmission rights and hedge structures before delivery, mainly through long-term rights governed by the Forward Capacity Allocation guideline (FCA). This layer transfers future price and congestion risk across zones but does not remove physical constraints; it defines admissible hedge translation and in turn constrains feasible spot decisions in later horizons.\cite{eu_reg_2016_1719,ec_network_codes_2026}

Day-ahead operation is executed in the Single Day-Ahead Coupling (SDAC) under shared coupling functions where order books from designated NEMOs are matched jointly with cross-zonal capacities. Clearing solves one coupled optimization problem across participating bidding zones, with algorithmic outcomes determined by both bid structure and transfer limits, not by local merit orders in isolation.\cite{acer_market_coupling_development_2026,acer_nemo_designation_2026,nord_pool_day_ahead_2026}

Intraday and balancing layers absorb forecast error and outage shocks under progressively tighter timing constraints. Intraday continuous trading updates the commercial position close to delivery, while balancing activation and settlement resolve the residual physical deviation that remains after market close. In system terms, balancing is the terminal correction stage of the same control process.\cite{nord_pool_intraday_2026,ec_network_codes_2026,eu_reg_2019_943}

Balancing design in the EU is structurally product- and platform-specific. The Electricity Balancing Guideline (EB GL) defines harmonization requirements for reserve products and activation processes, while cross-border activation is organized through dedicated platforms for automatic frequency restoration reserve (aFRR, via PICASSO), manual frequency restoration reserve (mFRR, via MARI), and replacement reserve processes linked to TERRE implementation history. These platform mechanics directly affect residual-risk transfer between intraday correction and post-delivery imbalance settlement.\cite{eu_reg_2017_2195_ebgl,entsoe_picasso_2026,entsoe_mari_2026,entsoe_terre_2026}

The design problem is not only algorithmic coupling but institutional heterogeneity under cross-border execution. ACER documents strong welfare gains from integration and, at the same time, persistent bottlenecks in capacity availability and implementation quality across jurisdictions. Combined with a multi-exchange topology under NEMO passporting, this produces real interface friction in data synchronization, gate-timing coordination, and compliance encoding for pan-European strategies.\cite{acer_final_assessment_report_2022,acer_market_design_assessment_2022,acer_nemo_list_jan_2026}

\section{Actors and rules}
The rule hierarchy combines sector regulation, market-coupling codes, and exchange-level rulebooks. Regulation (EU) 2019/943 sets internal-market principles and cross-border participation logic, while the code stack links allocation horizons and balancing obligations through the Capacity Allocation and Congestion Management regulation (CACM), FCA, and EB GL. The 2024 reform package kept marginal pricing and tightened long-term risk-allocation and consumer-protection instruments, which changes portfolio governance without removing the coupling architecture.\cite{eu_reg_2019_943,ec_network_codes_2026,eu_reg_2016_1719,ec_electricity_market_design_2026}

Integrity and surveillance obligations run in parallel to market-design obligations. REMIT governs wholesale market integrity and transparency, and operationally this means that trading organizations must separate strategy logic from prohibited-information use and manipulation patterns while preserving auditable decision records. Derivatives activity in financial venues adds conduct, transparency, clearing, and reporting constraints under MiFID II, MiFIR, and EMIR, so compliance control is not a post-trade layer; it is part of executable strategy design.\cite{eu_reg_2011_1227_remit,eu_dir_2014_65_mifid2,eu_reg_2014_600_mifir,eu_reg_2012_648_emir}

The spot exchange landscape is structurally plural. Under CACM designation rules, each Member State designates at least one NEMO and designated operators can passport services across borders under defined conditions. ACER's January 2026 list shows a multi-operator topology with major designated venues including EPEX Spot, Nord Pool, OMIE, GME, HEnEx, HUPX, OTE, IBEX, CROPEX, EXAA, TGE, OKTE, OPCOM, BRM, BSP SouthPool, SEMOpx, and ETPA across day-ahead and intraday service areas.\cite{acer_nemo_designation_2026,acer_nemo_list_jan_2026}

For implementation, exchange topology should be encoded as a structured venue matrix with operator and ownership context. \cref{tab:exchange-matrix} extends coverage to all major exchanges listed in the current NEMO landscape.\cite{acer_nemo_list_jan_2026}

\begin{table}[htbp]
\centering
\caption{Major EU short-term electricity exchanges and operator context compiled from ACER and operator disclosures.}
\label{tab:exchange-matrix}
\footnotesize
\setlength{\tabcolsep}{4pt}
\renewcommand{\arraystretch}{1.12}
\begin{tabularx}{\textwidth}{p{1.8cm} p{1.7cm} X p{2.6cm}}
\toprule
Exchange & Segment & Operator / ownership context & Link \\
\midrule
EPEX Spot & DA, ID & EPEX SPOT SE; part of EEX Group.\cite{epex_company_information_2026} & \href{https://www.epexspot.com/en/company-information}{epexspot.com} \\
Nord Pool & DA, ID & Nord Pool operator; acquired by Euronext.\cite{nord_pool_about_2026} & \href{https://www.nordpoolgroup.com/en/about-us/}{nordpoolgroup.com} \\
OMIE & DA, ID & Iberian operator under OMI Group structure.\cite{omie_about_2026,omi_group_faq_ownership_2026} & \href{https://www.omie.es/en/about-us}{omie.es} \\
GME & DA, ID & Italian operator GME; wholly owned by GSE.\cite{gme_organization_2026} & \href{https://www.mercatoelettrico.org/en-us/Home/Aboutus/Organization}{mercatoelettrico.org} \\
HEnEx & DA, ID & Hellenic Energy Exchange market operator.\cite{henex_company_profile_2026} & \href{https://www.enexgroup.gr/web/guest/company-profile}{enexgroup.gr} \\
HUPX & DA, ID & Hungarian exchange operator; ADEX group context.\cite{hupx_about_2026,adex_group_about_2026} & \href{https://hupx.hu/about}{hupx.hu} \\
OTE & DA, ID & Czech electricity and gas market operator.\cite{ote_basic_facts_2026} & \href{https://www.ote-cr.cz/en/basic-facts}{ote-cr.cz} \\
IBEX & DA, ID & Bulgarian exchange; sole owner is BSE.\cite{ibex_profile_2026} & \href{https://ibex.bg/en/profile.html}{ibex.bg} \\
CROPEX & DA, ID & Croatian exchange operator with system-institution ownership context.\cite{cropex_about_2026} & \href{https://www.cropex.hr/en/about-us}{cropex.hr} \\
EXAA & DA & Austrian exchange with published shareholder structure.\cite{exaa_about_2026} & \href{https://www.exaa.at/en/about-exaa/}{exaa.at} \\
TGE & DA, ID & Towarowa Gielda Energii in GPW Group perimeter.\cite{gpw_h1_2025_tge_ownership_2025} & \href{https://pap-mediaroom.pl/sites/default/files/2025-09/GPW_Group_H1_2025_Consolidated_Financial_Report.pdf}{GPW report} \\
OKTE & DA, ID & Slovak short-term market operator (OKTE, a.s.).\cite{okte_dam_overview_2026} & \href{https://www.okte.sk/en/short-term-market/published-information-of-dam/day-ahead-detailed-overview/}{okte.sk} \\
OPCOM & DA, ID & Romanian operator in Transelectrica group structure.\cite{transelectrica_opcom_2026} & \href{https://www.transelectrica.ro/web/tel/opcom}{transelectrica.ro} \\
BRM & DA, ID & Romanian Commodities Exchange electricity-market operator.\cite{brm_markets_2026} & \href{https://brm.ro/en/markets/}{brm.ro} \\
BSP SouthPool & DA, ID & Slovenian operator; majority stake held by ADEX Group.\cite{bsp_about_2026,adex_bsp_majority_2024} & \href{https://www.bsp-southpool.com/about-us/}{bsp-southpool.com} \\
SEMOpx & DA, ID & NEMO operator in Ireland and Northern Ireland; EirGrid/SONI joint structure.\cite{semopx_about_2026} & \href{https://www.semopx.com/about/}{semopx.com} \\
ETPA & ID & Dutch operator with NEMO license and XBID linkage.\cite{etpa_about_2026,etpa_nemo_license_2024} & \href{https://www.etpa.nl/en/about}{etpa.nl} \\
\bottomrule
\end{tabularx}
\end{table}

This venue map matters because cross-border strategies fail operationally at interface boundaries, not at conceptual market level. Distinct operator governance, product eligibility, and procedural timetables require explicit per-venue rule encoding even under harmonized coupling frameworks.\cite{acer_market_coupling_development_2026,acer_final_assessment_report_2022}

Cross-border execution is coordinated through coupled algorithms and synchronized gate windows, not by independent local clearing. In day-ahead coupling, matching incorporates cross-zonal network constraints and common algorithmic procedures; in intraday coupling and local continuous books, execution follows stricter timing and order-priority mechanics close to delivery. Nord Pool documents these mechanics explicitly through SDAC/Euphemia-based day-ahead matching and first-come first-served intraday execution with 15-minute, 30-minute, hourly, and block products.\cite{acer_market_coupling_development_2026,nord_pool_day_ahead_2026,nord_pool_intraday_2026}

The derivatives layer is institutionally distinct from the NEMO spot layer and remains critical for hedge transfer, collateral planning, and basis-risk management. EEX positions itself as the central European power derivatives venue with integrated spot affiliation through EPEX Spot, ICE Endex provides a large continental energy derivatives venue across gas, emissions, and power, and Euronext runs Nordic and Baltic power futures and EPAD contracts with the current post-migration structure launched in 2026.\cite{eex_power_2026,ice_endex_2026,euronext_nord_pool_power_2026}

For system architecture, the consequence is a permission matrix rather than a single market-access flag. Venue membership, product eligibility, gate-closure timing, balancing responsibility, clearing route, and surveillance obligations jointly determine whether an action is executable. The trading stack must encode those constraints at order-construction time, since infeasible or non-compliant actions cannot be repaired downstream without economic loss or regulatory exposure.\cite{acer_nemo_designation_2026,eu_reg_2011_1227_remit,eu_dir_2014_65_mifid2,eu_reg_2014_600_mifir,eu_reg_2012_648_emir}

\section{Trading workflow}
Execution starts with rolling portfolio-state construction rather than order placement. Forecast updates for load, generation, and asset availability are mapped against open forward and derivatives positions to compute residual volume risk by delivery period and bidding zone. Long-term transmission-right positions from the FCA layer constrain feasible hedge translation across zones and must be represented before day-ahead order construction.\cite{eu_reg_2016_1719,eex_power_2026,ice_endex_2026,euronext_nord_pool_power_2026}

Day-ahead positioning is submitted through designated NEMOs under coupled market operation. Matching in SDAC is executed with common coupling functions that combine order books and cross-zonal network capacities at gate closure, so local bid quality and interzonal transfer limits determine clearing jointly. The resulting schedule is a coupled allocation outcome, not an exchange-isolated merit-order result.\cite{acer_nemo_designation_2026,acer_market_coupling_development_2026,nord_pool_day_ahead_2026}

After day-ahead publication, intraday trading absorbs forecast error and operational events through repeated re-optimization. Continuous order matching close to delivery follows strict time-priority mechanics, and cross-border capacity updates propagate directly into executable opportunities. In this phase, execution latency and order-book depth become binding decision variables because correction windows narrow as delivery approaches.\cite{acer_market_coupling_development_2026,nord_pool_intraday_2026}

Residual deviations are resolved through balancing processes administered by TSOs under the balancing-guideline framework. Balance-responsible parties remain exposed to imbalance outcomes when commercial positions and physical delivery diverge, and settlement is performed after delivery on measured imbalance quantities and applicable pricing rules. Cross-border activation pathways now depend on reserve-product and platform allocation, especially for aFRR and mFRR activation through PICASSO and MARI structures, while replacement-reserve design remains under active transition pressure.\cite{ec_network_codes_2026,eu_reg_2019_943,eu_reg_2017_2195_ebgl,entsoe_picasso_2026,entsoe_mari_2026,entsoe_terre_2026}

Post-trade processing closes the loop through confirmation, clearing, reporting, and surveillance controls across venue types. Spot and cross-border wholesale activity remains subject to REMIT integrity and transparency obligations, while derivatives legs are shaped by MiFID II and MiFIR market-structure requirements together with EMIR clearing and reporting duties. A production system must reconcile these layers into one timestamped audit trail linking forecast state, order intent, execution outcome, and settlement exposure.\cite{eu_reg_2011_1227_remit,eu_dir_2014_65_mifid2,eu_reg_2014_600_mifir,eu_reg_2012_648_emir}

\section{Licensing and permissions}
Market entry in European power trading requires two permission stacks that cannot be merged into one checklist. The first stack defines whether an entity is legally allowed to operate in the relevant regulatory perimeter. The second stack defines whether that same entity can technically and contractually execute orders in specific markets. Treating these stacks as one item creates recurrent onboarding failures and compliance gaps at go-live.

\subsection{Legal and regulatory prerequisites}
At wholesale level, REMIT registration is a pre-trade condition for reportable wholesale energy transactions: the market participant must register with the competent national regulatory authority and receive an ACER code through CEREMP before entering into transactions.\cite{eu_reg_2011_1227_remit,acer_remit_registration_market_participants_2026} Where automated execution is used, the revised REMIT framework adds a specific algorithmic-trading perimeter with notification and identifier obligations under Article 5a implementation practice.\cite{eu_reg_2024_1106_remit2,acer_remit_algorithmic_notifications_2024} If the business model extends into power derivatives as financial instruments, the legal perimeter expands to MiFID II and MiFIR conduct and market-structure obligations together with EMIR clearing and reporting duties.\cite{eu_dir_2014_65_mifid2,eu_reg_2014_600_mifir,eu_reg_2012_648_emir} In parallel, physical delivery exposure remains tied to balancing responsibility under EU electricity-market rules, either via own BRP setup or a contracted balance-responsibility arrangement.\cite{eu_reg_2019_943,eu_reg_2017_2195_ebgl}

\subsection{System and venue permissions}
Legal status does not grant executable market access. Venue access is governed by exchange and clearing documentation, participant agreements, admission checks, and technical onboarding processes. Nord Pool documents this explicitly through rulebook and participant-agreement scope plus member-approval conditions that include authorization and fit-and-proper requirements.\cite{nord_pool_rules_regulations_2026,nord_pool_general_terms_conditions_2024} In the derivatives layer, EEX access documentation states the same separation structurally: exchange participation and clearing admission at ECC are distinct permissions that must both be in place before production trading.\cite{eex_access_trading_2026}

From a system-design perspective, this separation means a trading platform must encode two independent gates for every action candidate: a legal-regulatory gate and a venue-execution gate. The first gate validates entity-level regulatory admissibility; the second gate validates market-, product-, and counterparty-level execution permissions. Only their intersection defines executable actions in production.

\bigskip

\section{System requirements}
A decision-ready trading system requires a compact state representation that merges market microstructure, physical position, and regulatory feasibility. The minimum operational state is defined in Equation~\eqref{eq:state-vector} as a tuple of market descriptors, portfolio state, transfer limits, executable prices, risk state, and compliance state at decision time $t$.
\begin{equation}
\mathbf{s}_t = \left(\mu_t,\nu_t,\kappa_t,\pi_t,\rho_t,\sigma_t\right)
\label{eq:state-vector}
\end{equation}
In this notation, $\mu_t$ denotes market-status variables such as gate state and venue mode, $\nu_t$ the physical and contractual portfolio state, $\kappa_t$ cross-zonal capacity state, $\pi_t$ executable price and depth state, $\rho_t$ risk state, and $\sigma_t$ compliance state under market-integrity and venue rules.\cite{acer_market_coupling_development_2026,acer_nemo_designation_2026,ec_network_codes_2026,eu_reg_2011_1227_remit}

Residual delivery exposure must be computed per interval and bidding zone before any order is submitted. Equation~\eqref{eq:residual-exposure} defines the signed residual quantity that propagates from forecasting and hedge translation into day-ahead, intraday, and balancing decisions.
\begin{equation}
\xi_{t,z} = d_{t,z} - g_{t,z} - h_{t,z} - x^{\mathrm{DA}}_{t,z} - x^{\mathrm{ID}}_{t,z} - b_{t,z}
\label{eq:residual-exposure}
\end{equation}
Here $d_{t,z}$ is forecast demand, $g_{t,z}$ controllable generation, $h_{t,z}$ forward and derivatives hedge volume translated to zone $z$, $x^{\mathrm{DA}}_{t,z}$ and $x^{\mathrm{ID}}_{t,z}$ cleared day-ahead and intraday positions, and $b_{t,z}$ activated balancing volume. The residual $\xi_{t,z}$ is the direct driver of imbalance exposure.\cite{eu_reg_2016_1719,nord_pool_day_ahead_2026,nord_pool_intraday_2026,eu_reg_2019_943}

A production objective must combine execution cost, imbalance risk, and regulatory cost in one optimization problem. Equation~\eqref{eq:dispatch-objective} formalizes this as a constrained objective over market actions $\mathbf{x}$.
\begin{align}
\min_{\mathbf{x}} \quad
J &= \sum_{t,z}\left(\pi^{\mathrm{DA}}_{t,z}x^{\mathrm{DA}}_{t,z} + \pi^{\mathrm{ID}}_{t,z}x^{\mathrm{ID}}_{t,z} + \lambda_{t,z}\lvert \xi_{t,z}\rvert\right)
+ \alpha\,\mathrm{CVaR}_{\beta}(L) + \gamma C^{\mathrm{reg}}
\label{eq:dispatch-objective} \\
\text{s.t.}\quad
-\kappa_{t,\ell} &\le \phi_{t,\ell} \le \kappa_{t,\ell},\qquad
\tau \le \tau^{\mathrm{GC}}_{t,m} \nonumber
\end{align}
In Equation~\eqref{eq:dispatch-objective}, $\lambda_{t,z}$ prices imbalance risk, $\mathrm{CVaR}_{\beta}(L)$ is the tail-risk term at confidence level $\beta$, $\alpha$ and $\gamma$ are policy weights, $\phi_{t,\ell}$ is scheduled flow on interconnector $\ell$, $\kappa_{t,\ell}$ is available transfer capacity, and $\tau^{\mathrm{GC}}_{t,m}$ is the gate-closure time for market $m$. These constraints encode coupling and timing rules directly in the optimizer.\cite{acer_market_coupling_development_2026,nord_pool_day_ahead_2026,ec_network_codes_2026}

Execution feasibility cannot be inferred from price signals alone; it must be checked against venue permissions and regulatory predicates at order granularity. The executable-action indicator in Equation~\eqref{eq:permission-gate} represents this gate.
\begin{equation}
\Omega_i =
\mathbf{1}\!\left[\tau_i \le \tau^{\mathrm{GC}}_{m(i)}\right]
\mathbf{1}\!\left[u_i \in \mathcal{U}_{m(i)}\right]
\mathbf{1}\!\left[p_i \in \mathcal{P}_{m(i)}\right]
\mathbf{1}\!\left[r_i \in \mathcal{R}_{m(i)}\right]
\label{eq:permission-gate}
\end{equation}
For order candidate $i$, $\Omega_i=1$ is required for release to the venue; $\mathcal{U}_{m(i)}$ is the authorized participant set, $\mathcal{P}_{m(i)}$ the permitted product set, and $\mathcal{R}_{m(i)}$ the active regulatory predicate set. This structure links exchange access, product scope, and surveillance obligations to the same control decision.\cite{acer_nemo_designation_2026,eu_reg_2011_1227_remit,eu_dir_2014_65_mifid2,eu_reg_2014_600_mifir,eu_reg_2012_648_emir}

\section{Battery storage and temporal arbitrage}
Merit-order pricing in markets with high renewable penetration produces structural temporal spreads. When wind or solar output exceeds demand, zero-marginal-cost generation depresses clearing prices; when output drops or demand peaks, scarcity pricing takes over. These are not transient anomalies but persistent features of the energy transition: as renewable capacity grows, both the frequency and magnitude of intra-day price swings increase. European intraday markets regularly exhibit spreads exceeding \qty{100}{\text{EUR/MWh}} between off-peak surplus hours and peak scarcity hours within the same delivery day.\cite{nord_pool_intraday_2026,nord_pool_day_ahead_2026}

Battery storage is the physical instrument that captures temporal spreads. The operating principle is direct: charge when the clearing price is low, discharge when it is high, and retain the difference minus round-trip efficiency losses and degradation cost. A storage operator with a \qty{100}{\mega\watt} / \qty{200}{\mega\watt\hour} system charging at \qty{20}{\text{EUR/MWh}} and discharging at \qty{140}{\text{EUR/MWh}} captures a gross margin of \qty{24000}{\text{EUR}} per cycle before efficiency, grid fees, and wear adjustments. Multiple cycles per day are feasible in markets with strong solar midday troughs and evening demand peaks, which compounds daily revenue.

The business case strengthens when a battery participates across multiple market layers simultaneously. Day-ahead arbitrage captures the base spread from overnight positioning. Intraday re-optimization captures additional value from forecast errors: when wind or solar output deviates from the day-ahead forecast, intraday prices shift and the battery adjusts its schedule. Balancing and frequency response services add a third revenue layer, where the battery earns activation payments for fast power injection or withdrawal under TSO procurement.\cite{eu_reg_2017_2195_ebgl,entsoe_picasso_2026,entsoe_mari_2026} This combination of revenue streams across horizons is the core of what practitioners call value stacking, and it is the reason that battery returns can exceed single-market projections by a wide margin.

From an architecture perspective, batteries represent the canonical multi-horizon optimization problem. Optimal dispatch requires joint consideration of day-ahead price expectations, intraday correction opportunities, balancing activation probabilities, state-of-charge constraints, and degradation cost across overlapping delivery windows. Each decision changes the feasible set for subsequent decisions: a battery that commits capacity to frequency response cannot simultaneously use that capacity for intraday arbitrage. This coupling between horizons and markets is precisely the problem that the state vector in Equation~\eqref{eq:state-vector} and the constrained objective in Equation~\eqref{eq:dispatch-objective} are designed to represent, making battery dispatch a primary validation case for the proposed architecture.

\section{AI architecture}
The architecture is modeled as a constrained policy system that maps operational state $\mathbf{s}_t$ to executable actions while preserving risk and regulatory bounds. Equation~\eqref{eq:ai-policy-objective} defines the optimization target as expected utility with explicit tail-risk and governance penalties.
\begin{align}
\max_{\theta}\quad
\mathcal{J}(\theta) &= \mathbb{E}\!\left[\sum_{t=1}^{T}\Big(r_t\!\left(a_t,\mathbf{s}_t\right)
- \alpha\,\mathrm{CVaR}_{\beta}(L_t)
- \gamma\,C_t^{\mathrm{gov}}\Big)\right]
\label{eq:ai-policy-objective}\\
\text{s.t.}\quad
\Omega_t(a_t) &= 1,\qquad a_t \in \mathcal{A}(\mathbf{s}_t) \nonumber
\end{align}
In Equation~\eqref{eq:ai-policy-objective}, $\theta$ parameterizes the decision policy, $r_t$ is realized economic reward, $\mathrm{CVaR}_{\beta}(L_t)$ is the tail-loss term at confidence level $\beta$, and $C_t^{\mathrm{gov}}$ represents governance and control cost. The executable-action constraint $\Omega_t(a_t)=1$ links the policy directly to venue and rule feasibility.\cite{rockafellar_uryasev_2000_cvar,eu_reg_2011_1227_remit,eu_dir_2014_65_mifid2,eu_reg_2014_600_mifir,eu_reg_2012_648_emir}

Forecasting and opportunity ranking use calibrated predictive distributions rather than point forecasts. Equation~\eqref{eq:conformal-coverage} specifies a distribution-free coverage requirement for the predictive set $\Gamma_{1-\delta}(\mathbf{s}_t)$, which is consumed by downstream decision and risk modules.
\begin{equation}
\mathbb{P}\!\left(y_{t+h}\in \Gamma_{1-\delta}(\mathbf{s}_t)\right)\ge 1-\delta
\label{eq:conformal-coverage}
\end{equation}
The confidence parameter $\delta$ controls interval width and directly influences aggressiveness in intraday repositioning; lower $\delta$ increases protection against forecast misspecification at the cost of reduced expected capture.\cite{angelopoulos_bates_2021_conformal,nord_pool_intraday_2026}

Execution control is fail-closed by construction. Equation~\eqref{eq:ai-fail-closed} enforces release only when profitability, risk, and compliance predicates are simultaneously satisfied.
\begin{equation}
a_t =
\begin{cases}
\arg\max\limits_{a\in\mathcal{A}(\mathbf{s}_t)} U_t(a), & \Omega_t(a)=1\ \land\ R_t(a)\le \bar{\rho}\ \land\ M_t(a)\ge \underline{\mu}\\
\varnothing, & \text{otherwise}
\end{cases}
\label{eq:ai-fail-closed}
\end{equation}
Here $U_t(a)$ is utility, $R_t(a)$ predicted risk, $\bar{\rho}$ the risk budget, $M_t(a)$ a market-quality predicate combining depth and latency conditions, and $\underline{\mu}$ its acceptance threshold. The null action $\varnothing$ is a valid output and preserves safety when constraints are violated.\cite{acer_market_coupling_development_2026,eu_reg_2011_1227_remit}

Operational robustness requires explicit model-risk supervision and automatic strategy downgrades. Equation~\eqref{eq:ai-override-rule} defines the override controller that switches to a conservative policy when either distribution drift or calibration error exceeds control limits.
\begin{equation}
\pi_t =
\begin{cases}
\pi_{\theta}, & \Delta_t \le \eta_{\Delta}\ \land\ \varepsilon_t \le \eta_{\varepsilon}\\
\pi_{\mathrm{safe}}, & \text{otherwise}
\end{cases}
\label{eq:ai-override-rule}
\end{equation}
The drift statistic $\Delta_t$ tracks state-distribution shift, $\varepsilon_t$ measures uncertainty-calibration error, and $\eta_{\Delta},\eta_{\varepsilon}$ are supervisory thresholds tuned to tolerated operational risk.\cite{nist_ai_rmf_2023,ec_network_codes_2026}

Auditability is treated as a first-class output variable. Equation~\eqref{eq:audit-record} defines the minimal immutable record tuple stored for each decision.
\begin{equation}
\ell_t=\left(\mathbf{s}_t,\hat{y}_t,\Gamma_{1-\delta}(\mathbf{s}_t),a_t,\Omega_t,\pi_t,\xi_t\right)
\label{eq:audit-record}
\end{equation}
The record $\ell_t$ captures state, forecast, uncertainty set, executed action, feasibility gate result, active policy, and residual exposure $\xi_t$, which allows line-by-line reconstruction from forecast input to settlement consequence. This structure supports surveillance expectations under REMIT and documentation discipline aligned with EU AI governance requirements.\cite{eu_reg_2011_1227_remit,eu_reg_2024_1689_ai_act}

A reproducible validation protocol requires a fixed metric vector spanning economic, risk, compliance, and latency dimensions. Equation~\eqref{eq:validation-vector} defines the minimal validation tuple for model and policy comparison.
\begin{equation}
\mathbf{v}=\left(\mathbb{E}[\Pi],\mathrm{CVaR}_{\beta}(L),C^{\mathrm{imb}},\mathrm{FPR}^{\mathrm{comp}},\mathrm{FNR}^{\mathrm{comp}},\Lambda^{95}\right)
\label{eq:validation-vector}
\end{equation}
Here $\mathbb{E}[\Pi]$ is expected trading result, $\mathrm{CVaR}_{\beta}(L)$ tail-loss risk, $C^{\mathrm{imb}}$ imbalance settlement cost, $\mathrm{FPR}^{\mathrm{comp}}$ and $\mathrm{FNR}^{\mathrm{comp}}$ compliance false-positive and false-negative rates, and $\Lambda^{95}$ the 95th-percentile decision-to-order latency.

Backtesting should be executed over an explicit stress set rather than a single historical regime. Equation~\eqref{eq:stress-set} defines a compact scenario basis for evaluation across normal operation, congestion stress, forecast shock, outage regimes, and decoupling fallback events.
\begin{equation}
\mathcal{S}=\left\{s^{\mathrm{base}},s^{\mathrm{cong}},s^{\mathrm{forecast}},s^{\mathrm{outage}},s^{\mathrm{decouple}}\right\}
\label{eq:stress-set}
\end{equation}
For each $s\in\mathcal{S}$, the policy is accepted only if performance improves on baseline while preserving feasibility-gate integrity and governance thresholds. This design aligns risk validation with AI lifecycle supervision and platform-level market constraints.\cite{nist_ai_rmf_2023,rockafellar_uryasev_2000_cvar,acer_market_coupling_development_2026}

\section{Discussion}
Forecasting reviews benchmark model accuracy for day-ahead, intraday, and balancing prices, while reinforcement-learning studies optimize bid construction for single-venue auction settings.\cite{lago_marcjasz_schutter_weron_2021_epf,oconnor_bahloul_prestwich_visentin_2025_epf,di_persio_garbelli_giordano_2025_rl_bidding} These contributions improve local component performance but treat market layers as independent optimization contexts. The architecture proposed here differs by modeling horizon coupling, cross-zonal capacity constraints, and multi-venue permission rules as joint inputs to a single decision process.\cite{acer_market_coupling_development_2026,eu_reg_2019_943} Under this framing, a day-ahead bid is not an isolated optimization output; it is a coupled decision whose feasibility depends on forward hedge state, intraday correction capacity, and balancing-layer exposure.

The permission gate formalized in Equation~\eqref{eq:permission-gate} shifts compliance enforcement from post-trade monitoring to pre-trade construction. In conventional system design, regulatory checks are applied as filters after an optimizer produces candidate orders. The architecture inverts this relationship: venue membership, product eligibility, gate-closure timing, and regulatory predicates enter the optimization as hard constraints, so infeasible or non-compliant actions cannot be generated.\cite{acer_nemo_designation_2026,eu_reg_2011_1227_remit} The fail-closed logic in Equation~\eqref{eq:ai-fail-closed} extends this principle to the AI decision layer, where the system defaults to inaction when any constraint is violated rather than requiring explicit exception handling for each failure mode.

The EU AI Act and the NIST AI Risk Management Framework define lifecycle governance structures for AI systems but do not specify how these structures map to sector-specific market mechanics.\cite{eu_reg_2024_1689_ai_act,nist_ai_rmf_2023} The architecture proposed here operationalizes governance for electricity trading by embedding auditability, uncertainty calibration, and override controls as system components rather than documentation requirements. The audit record tuple in Equation~\eqref{eq:audit-record} and the drift-triggered policy override in Equation~\eqref{eq:ai-override-rule} translate abstract governance expectations into executable logic that can be tested, monitored, and verified against regulatory thresholds.

\section{Conclusion}
The paper proposes a unified interpretation of European electricity trading as a coupled control problem across forward, day-ahead, intraday, and balancing horizons. Market outcomes are shown as joint products of capacity-constrained coupling, venue timing rules, and actor-specific obligations rather than isolated auction results.\cite{eu_reg_2019_943,ec_network_codes_2026,acer_market_coupling_development_2026}

The governance analysis indicates that executable trading logic is defined by an interaction of sector and financial regulation with exchange-level permissions. REMIT, MiFID II, MiFIR, and EMIR map directly into pre-trade admissibility checks and post-trade accountability requirements, while NEMO designation and passporting rules explain the persistent multi-venue topology of EU spot markets.\cite{eu_reg_2011_1227_remit,eu_dir_2014_65_mifid2,eu_reg_2014_600_mifir,eu_reg_2012_648_emir,acer_nemo_designation_2026,acer_nemo_list_jan_2026}

The system contribution is a minimal decision architecture that links state representation, residual exposure accounting, constrained optimization, uncertainty calibration, and fail-closed execution control. This framework proposes how AI support can be integrated without separating economic optimization from regulatory feasibility or auditability.\cite{rockafellar_uryasev_2000_cvar,angelopoulos_bates_2021_conformal,nist_ai_rmf_2023,eu_reg_2024_1689_ai_act}

The resulting position is methodological and operational: model the market as one constrained system, encode permissions before execution, and treat uncertainty and governance as first-class state variables. Under this framing, AI is proposed as a bounded decision component inside regulated market operation, not as an autonomous layer outside institutional control.

\section{Limitations}
The framework is scoped to EU wholesale electricity trading and does not claim completeness for all national implementation detail. Code-level harmonization coexists with local market design differences, and the practical burden of rule interpretation remains jurisdiction-sensitive at delivery-zone and venue levels.\cite{acer_nemo_list_jan_2026,ec_network_codes_2026}

The operational model abstracts from data-path fragility that materially affects live performance: feed latency, revision frequency asymmetry, and outage behavior across venues and operators. These factors can dominate theoretical edge in intraday correction windows and can degrade any policy calibrated on clean historical snapshots.\cite{nord_pool_intraday_2026,acer_market_coupling_development_2026}

The legal environment is dynamic. Regulatory obligations in AI governance, market integrity, and financial-market supervision continue to evolve, which limits the temporal stability of any fixed compliance encoding. Production deployment requires continuous legal versioning and rule-to-control trace updates, not one-time policy certification.\cite{eu_reg_2024_1689_ai_act,eu_reg_2011_1227_remit,eu_dir_2014_65_mifid2,eu_reg_2014_600_mifir}

The quantitative formulation still requires empirical validation on synchronized multi-venue event streams with realistic transaction-cost, latency, and imbalance-penalty models. The next research stage is out-of-sample evaluation under stress regimes and controlled ablations of risk, calibration, and governance gates to estimate the marginal value of each architectural constraint.\cite{rockafellar_uryasev_2000_cvar,angelopoulos_bates_2021_conformal,nist_ai_rmf_2023}

\bigskip

\printbibliography

\swDeclarationsPage

\end{document}